# Towards Turkish ASR: Anatomy of a rule-based Turkish g2p

Duygu Altınok

ABSTRACT. This paper describes the architecture and implementation of a rule-based grapheme to phoneme converter for Turkish. The system accepts surface form as input, outputs SAMPA mapping of the all parallel pronounciations according to the morphological analysis together with stress positions. The system has been implemented in Python.

## 1.Introduction

In this paper, we represent design and implementation of a rule-based grapheme to phoneme conversion system for Turkish. Turkish has a very large, possibly infinite vocabulary due to productive reflectional and derivational morphology. Nouns have hundreds of forms and a verbal root can generate thousands of words. Hence our g2p is based on recognizing and processing the root of a given word.
SAMPA mappings of the words based on dictionary lookup for roots with exceptioary phonetic events and rule-based generation for regular roots. Stress marking comes from morphological analysis of the input word.
Turkish looks more or less a phonetic language at the first glance, but old Arabic, Persian loan words break both pronunciation and stress rules. Abbreviation and foreign word handling is a major issue that we put effort in.
This paper is not the first Turkish pronunciation lexicon. Oflazer and Inkelas already implemented a finite state lexicon for Turkish in [11] with extensive use of finite state tools. What we introduce diffent from this extensive paper is, intensive effort on abbreviations and foreign words.
We implemented whole system in Python. Implementation details and corresponding GitHub repository is also provided.

## 2. Turkish

### 2.1 Turkish morphology

Turkish is an Ural-Altaic language, having agglutinative word structures. Turkish has a productive morphology with inflectional and derivational processes. Theoreticaly, there are infinite number of words in Turkish as one root may generate hundreds of new reflected, derived words. For instance take the verbal root *kazmak* and see its reflections: *kazdın, kazdım, kazdı, kazdık, kazdılar, kazdınız, kazarken, kazıyorken, kazmazken, kazmıyorken, kazmazdı, kazmazdım…*
Turkish morphology has been well studied by Oflazer [13]. A two-level morphological analyzer for Turkish was built by using XRCE finite state tools. This analyzer segments words into a series of lexical morphemes.
Because of complex morphophonological processes, we'll annotate classes of graphemes e.g. *A* represents unrounded back vowels *a* and *e*, *H* is for high vowels *ı, i, u, ü* and *D* stands for *d* and *t*.

### 2.2. Turkish phonology

Turkish ortography has 8 vowels *a, e, ı, i, o, ö, u, ü*; corresponds to /a, e, 1, i, o, 2, u, y/. Every vowel has a long counterpart, and represented with an additional *:* in SAMPA;  /a:, e:, 1:, i:, o:, 2:, u:, y:/. /o:/ and /2:/ are not very common, indeed these phonemes doesn't exist in words but exists in abbreviations. Vowels are symmetric upto backness, roundness, and height. Vowels harmonize within a root, and suffixes harmonize with the last vowel of the root.
Turkish has 26 consonants /p, t, tS, k, c, b, d, dZ, g, gj, f, s, S, v, w, z, Z, m, n, N, l, 5, r, j, h, G/. /G/ known as soft g, is a controversial consonant of Turkish. Please see the soft g section. From rest of the consonants, only *k, l* and *g* are tricky as they have 2 allophones;  /c, k/, /l, 5/ and /gj, g/; palatal and nonpalatal allophones of these graphemes. Palatalized segments occur in the existence of front vowels and nonpalatal counerparts occurs with back vowels. Except some roots, palatal/nonpalatal occurances are predictable because of frontness – backness, hence can be turned into rules. Overviews of Turkish phonology can be found in [2], [3] and [9].

**2.3. Turkish stress pattern**

Turkish stress pattern is highly morphology dependent. Most of the native words carry stress in their last syllables, usually suffixes shifts the stress "towards" end of the word. Word-final stress is more or less standart for most words, exceptions occur with old Arabic/Persian loan words.
Geographical names has a different stress pattern, known as Sezer stress [14]. Compound words are also different, first word keeps its stress and second word loses ; hence stress is again nonfinal.
Our stress computations are rule-based as well. We considered regular roots, irregular roots, compound words and geographical names with regular and stress shifting suffixes. Please see the corresponding section for the implementation details.

**3. Architecture of the g2p system**

Our system consists of composition of several modules. We first make morphological analysis of the surface,. Then basicly we need to get the root pronunciation and concetenate it to the suffix sequence pronunciation. All of the interesting phonological events happens at either root itself or on the root boundary; pronunciation of the suffix sequence is quite straightforward. Hence once we recognize the root, rest is easy. Phontecizier module contains a root pronunciation generator for native Turkish roots, one for foreign words and one for abbreviations together with a root and suffix sequence pronunciation combiner.
When all possible SAMPA mappings of the word are ready, we can check for some exceptionary phonological events that happens in speech but not reflected to ortography. Finally we produce stress markers and syllabified form for TTS mode. We don't need to produce stress positions and syllabification for the ASR, see below.

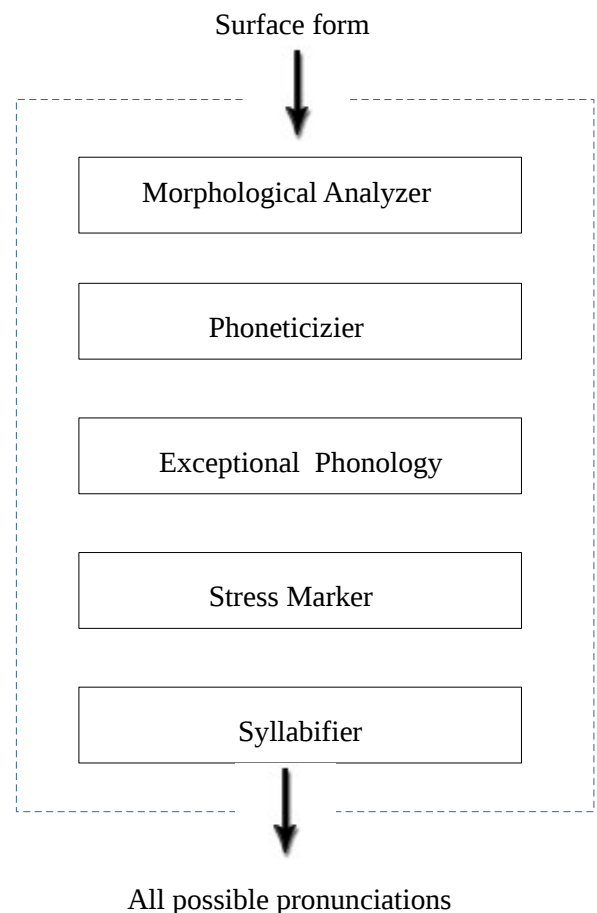

Fig.1. Module structure

### 3.1. Morphological Analyzer

We used Çağrı Çöltekin's TRmorph. In [1], he described his work as "To our knowledge, TRmorph is the first freely available morphological analyzer for Turkish". TRmorph provides stemming and full morphological analysis functionalities for Turkish. We preferred TRmorph for its great variety of functionalities.

Our morphological analyzer module consists of two parts, TRmorph's full morphological analyzer for ordinary words and our heuristic stemmer for loan, foreign words and abbreviations. See 4.1. for heuristic stemming.

Input to this module is the surface form of the word and output is a list of the tuples of the form (root, suffix sequence, analysis). For instance word *koyun*:

*koyun* → [(*koy*, *un*, <V><imp><2p>), (*koyu*, *n*, <Adj><0><N><p2s>), (*koy*, *un*, <N><gen>), (*koyun*, , <N>)]

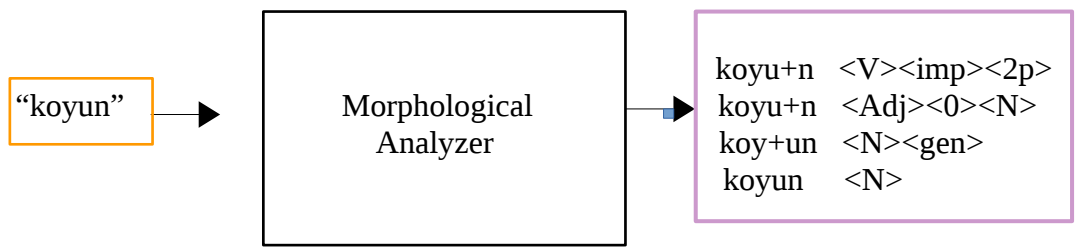

Fig.2. Output of the morpholical analyzer

As seen output of this module is set of possible roots and corresponding analysis of the surface form, to be fed to the phoneticizier.

### 3.2. Phoneticizier

Our phoneticizier module consists of three submodules. First is a rule-based SAMPA mapper for the words that we know to be Turkish. Second SAMPA mapper deals with foreign, loan words and abbreviations. For the latter part please see the next section. Third submodule unites root and suffix sequence pronunciations. First introduce the rule-based SAMPA mapper.

We hold a dictionary and 5 lists related to this module. First dictionary is root pronunciations. If a root has long vowels or irregular phonetics, this root definitely goes to this dictionary. Many roots are just 1-to-1 mappings of its graphemes. Many roots are old loan Arabic/Persian words and has long vowels. For instance *abide* is an old Arabic loan word and pronunciation is /a: b i d e/, definitely not 1-to-1 its graphemes. A second example is a native Turkish word *çiftlik*. Sometimes t is not pronounced, hence this word has 2 possible sayings: /tS i f t l i c/ and /tS i f l i c/. Words doesn't follow phonology rules also goes to this dictionary. Word *lale* doesn't follow palatalization rules of Turkish. *l* is realized as /5/ within the same syllable with a front vowel in native words. *lale* is pronounced as /l a l e/, first *l* doesn't follow the rule hence this word gets into the dictionary with its pronunciation.

Second type of irregular words have vowel length alterations in the case of reflection. For instance root *zaman* /z a m a n/ has no long vowels. In the case of reflecting with a suffix beginning with a vowel last syllable of the root lengthens:

*zaman+a* /z a m a: n a/
*zaman+ında* /z a m a: n 1 n d a/ .

If the suffix begins with a consonant basicly nothing happens:

*zamanda* /z a m a n d a/
*zamanla* /z a m a n 5 a/ .

This sort of irregular words goes to the irregular roots list with a special irregularity marker.
Another type of roots that goes into this list is the ones with final consonant voicing irregularity. Most of the Turkish roots ending with a voiceless consonants p, ç, t and k this final consonant changes to its voiced counterpart before a suffix beginning with a vowel:

*kitap+a       kitaba*
*ceset+e       cesede*
*çekiç         çekici*

Some roots don't follow this rule. Mostly one syllable words, old loan words and foreign words comes into this category.

*cumhuriyet+e   cumhuriyete*
*tank+ı         tankı*

From the view of g2p, there is absolutely no problem. Problem is with proper nouns. Proper nouns accepts reflectional suffixes with an apostrophe and even if final voicing occurs, it's not shown in orthography:

| Proper Noun | Surface | Pronunciation | SAMPA |
|---|---|---|---|
| *Zongulda**k**+a* | *Zongulda**k**'a* | *Zongulda**ğ**a* | /z o n gj u 5 d a **G** a/ |
| *Ahme**t**+e* | *Ahme**t**'e* | *Ahme**d**e* | /a h m e **d** e/ |

Some of the proper nouns doesn't go through final consonant voicing:

| Proper Noun | Surface | Pronunciation | SAMPA |
|---|---|---|---|
| *Sar**p**+a* | *Sar**p**'a* | *Sar**p**a* | /s a r **p** a/ |
| *Mins**k**+e* | *Mins**k**'e* | *Mins**k**e* | /m i n s **c** e/ |

Hence we need to know which proper noun roots follow final voicing rule. This sort of irregular roots gets into the irregular root list with a special "doesn't soften" mark.
We have 4 more lists, proper nouns list, geographical names list, abbreviations list and foreign words list, we call them genre lists.

Input to word-to-SAMPA comes from morphological analyzer as (root, suffix sequence). First we check which genre list the root is in. If root is an abbreviation or a foreign word, we feed the (root, suffix, sequence) tuple to the abbreviation, loan/foreign word module. Else, we feed the analysis tuple to the rule-based phoneticizier with genre info; geographical name, proper noun or ordinary word.

All SAMPA mappers first lookup the root in corresponding dictionaries. If exists, we have the root pronunciations. Note that there might be more than 1 pronunciation. Else, we make a SAMPA mapping according to the Turkish rules. Foreign word and abbreviations are explained in the next section. We'll focus on the ordinary word mapper and explain some of the Turkish phoneticization rules. First rule is about allophones of *k*, *l* and *g*. *k*, *l* and *g* has palatalized /c/, /l/, /gj/ and nonpalatalized /k/, /5/, /g/ allophones. Palatalized ones occurs in the existence of front vowels in the same syllable. Second rule is about loan roots that begin with two consonants e.g. *prenses, kral, tren*. We place a vowel according to the first vowel in the word:

```
kral    k 1 r a 5
prens   p i r e n s
gram    g 1 r a m
grup    g u r u p
```

That's it. Either we look the root pronunciation up in dictionary or we make a rule-based mapping. After having the root pronunciation, we generate the suffix sequence pronunciation. We again make a rule-based mapping, computationally less interesting. "Interesting" phonological phenomena occurs either in root itself or in the stem boundary.

Now we have root pronunciation either from foreign/loan/abbreviation phoneticizier or native word phoneticizier; also suffix sequence pronunciation. We need to unite the root and suffix sequence pronunciations. Here, we need to be careful. There are several considerations we should take into account:
- Irregular vowel alterations during refelction, word like zaman /z a m a n/ and zaman+a /z a m a: n a/.
- Proper nouns which doesn't follow final consonant devoicing rules e.g.

We make a irregular roots list look up, check if the root has any of the irregularities and proceed accordingly.

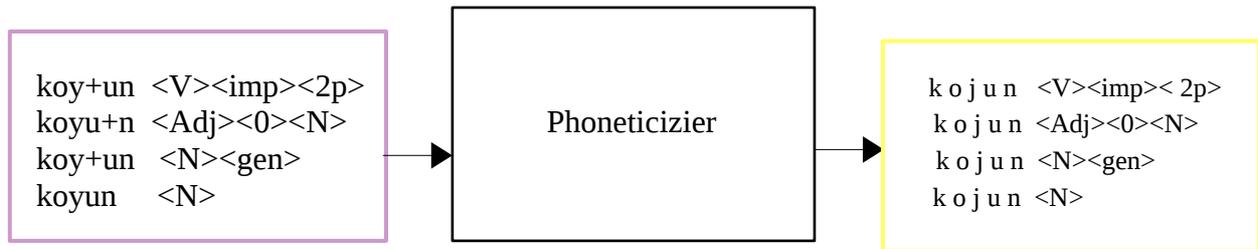

Fig.3. Output of the phoneticizier

### 3.3. Exceptional Phonology Module

Exceptional phonology module computes phonological phenomena that happens in spoken language but not represented in written language. We implemented several methods:

#### 3.3.1. n-1 assimilation

*l* generally gets assimilated by a preceding *n* in fast speech e.g. *kadınlar* may be pronounced /k a d 1 n n a r/ and *anla-* may be pronounced /a n n a-/.

#### 3.3.2. Rounding *r* in present tense suffix *-yor*

Most native speakers cut *r* sound in present tense suffix *-yor* if the suffix is followed by a suffix begins with a consonant or *-yor* is the last suffix of the word itself. Compare

*gidiyor*      /gj i d i **j o**/  
*gidiyordum*   /gj i d i **j o** d u m/  
*gidiyorken*   /gj i d i **j o** c e n/  

to

*gidiyorum*    /gj i d i **j o r** u m/  
*gidiyoruz*    /gj i d i **j o r** u z/ .

This method requires to know the morphological analysis of the word, since we're interested in *yor* only if it's the present tense suffix. Some roots contains *yor* as substrings e.g. *yorum, yorul-, yortu*, therefore morphological analysis of the word is needed here.

### 3.3.3. Realizations of *ğ*

*ğ* is known as "soft g" and is the ninth letter of the Turkish alphabet. Although classified as consonant, soft g lack a corresponding consonantal sound. It provides a smooth transition between consecutive vowels in loan words. Most native speakers have a glottal stop in consecutive vowels of old Arabic loan words like in *saat* /s a: a t/. Glottal stop is noticed not only in old loan words, but also new loan words and foreign words, as native Turkish doesn't have vowel clusters. For instance Maria is pronounced as /m a r i j a/. See the following section for details.

Turkish SAMPA has phoneme for soft g, /G/ although we didn't frequently used it. We rather tried to give soft g effect onto the surrounding vowels.

From computational point of view, soft g is either inaudible as a consonant or may be pronounced as a palatal glide in the environment of front vowels and as a bilabial glide in the environment of rounded vowels. In particular:

- When it is in word-final or syllable-final position, it lengthens a preceding
back vowel e.g. *dağdan* /d a: d a n/ and *dağ* /d a:/.
- Between identical back vowels it is inaudible e.g. *uğur* /u: r/, *ağarmak* /a: r m a k/ and *sığır* /s ı: r/.
- Between identical front vowels it is either inaudible (sevdiğim [sevdi:m] 'that
- Between identical front vowels it is either inaudible e.g. *bildiğim* /b i l d i: m/, or sounds like a palatal glide e.g. *düğün* /d y j y n/.
- When it occurs between an *e* and an *i* it is either inaudible or pronounced as a palatal glide /j/. Word *değil* is often heard as /d e j i l/ and /d i: l/.
- In the occurrence of between *i* and *e*, soft g mostly heard as palatal glide /j/ : *diğer* is often pronounced as /d i j e r/ and incorrectly written as *diyer*.
- Between rounded vowels it is mostly inaudible e.g. *soğuk* /s o u k/.
- Between a rounded vowel and an unrounded vowel it is mostly inaudible e.g. doğan /d o a n/.
- *a+ğ+ı* sequences may either sound like a sequence of /a/ followed by /1/ or like a sequence of two /a/ vowels: *ağır* as /a 1 r/ or /a: r/.
- ı+ğ+a sequences are pronounced as sequences of /a/ followed by /1/ : *sığan* /s 1 a n/.

Please see the example lexicon file for examples.

### 3.3.4. Other common exceptional phenomena

*y* is the palatal glide. *i+y* sequence sometimes can be realized as /i:/ like in *iyi* /i:/ and *diyeceğim* /d i: dZ e m/.

Future tense suffix is usually shortened during fast speech and causes interesting phonological phennomena hence deserves our attention. Although words are written "properly" in written language, short forms are also written in colloquial written language. See the examples:

| Word | Shorter & Incorrect writing | Pronounced as |
|---|---|---|
| gideceğim | gidicem | /gj i d i c e m/ |
|  | gitçem | /gj i t tS e m/ |
| gidecekmiş | gidicekmiş | /gj i dZ I dZ e c m i S/ |
|  | gitçekmiş | /gj i t tS e c m i S/ |
| dönecek | döncek | /d 2 n dZ e c/ |
| atacak | atcak | /a t dZ a k/ |
|  | atçak | /a t tS a k/ |
| bulacak | bulcak | /b u 5 dZ a k/ |

Please see the example lexicon file for examples.

## 3.4. Syllabifier

We syllabified the final output of the SAMPA mapper for TTS purposes. All syllables in Turkish words have the same duration during pronunciation,

Turkish has a very regular syllable pattern. We present an algorithm for syllabificiation:

Find the position of the first vowel, call it pos.
If the pos+1 th phoneme is a vowel split the syllable from right to the pos.
Else
  If the pos+2 th phoneme is a vowel split the syllable from right to the pos.
  Else
    If the pos+3 th phoneme is a vowel split the syllable from right to the pos+1 th position.
    Else
      If pos+1, pos+2, pos+3 th consonants are /str/, /ktr/, /ctr/ or /ntr/ split the syllable right to the pos+1 th position.
      Else split the syllable from right to the pos+2 th position.

We syllabify the final output for TTS purposes. We put – between syllables.

## 3.5. Stress Computations

We produced primary stress for TTS purposes. Neural networks can learn Turkish stress pattern as stress pattern is also rule-based provided there's enough speech corpus; hence for ASR purposes no stress computation is necessary indeed.
We showed primary stress with "+" marker in the lexicon:
okuma    /o k +u m a/
          / o k u m +a/.
Stress computations are also rule-based. There are three classes of the words one needs to be careful: compound names, geographical names and rest of the words. Proper nouns also falls into the third category. Geopraphical names carries stress either in the -2 or -3 the syllable, depending on which one is closed (ends with a consonant). If a geographical name has 2 syllables, stress is in the first sylable. Reflecting or deriving a geographical noun doesn't change the stress position.
Compound words also have a different stres pattern from ordinary words. The first compound element retains its stress (prior to compounding) while the second element loses its stress. Reflecting or deriving a compound root doesn't change the stress position.
Ordınary words carry stress at their final syllables. Most of the suffixes doesn't change stress pattern, derived/reflected words of ordinary roots have stress at the final syllable as well. In the case of concetenating a stress shifting suffix, stress shifts to vowel to the left of the suffix. *yor*, *ki* and several stress shifting suffixes can be found in [4], also see the implementation.
There are also orindary roots with exceptionary stress positions. We tag them in our list of exceptionary roots.
Basicly the rule is, if the root carries stress in its last syllable, then adding suffixes shifts the stress to the last syllable of the derived/reflected word. Except if the suffix is a stress shifting suffix, stress "can't advance" to the right of this suffix. Else, if the root has a exceptional stress position for any reason, reflecting or deriving doesn't change the stress position. See the examples:

Geographical name

Ankara        +a n k a r a
Ankara'da    +a n k a r a d a
Mudanya      m u d +a n j a
Mudanya'dan  m u d +a n j a d a n
Aydın         +a j d 1 n

Regular ordinary word

aydın      a j d +1 n
aydından   a j d 1 n d +a n
koyun      k o j +u n
koyundan   k o j u n d +a n
kısa       k 1 s +a
kısadan    k 1 s a d +a n

Regular ordinary word with irregular suffixes

gidi**yor**ken   gj i d +i j o r c e n
kısadır          k 1 s a d +1 r

İrregular ordinary words and compounds

papatya     p a p +a t j a
papatyadan  p a p +a t j a d a n
bugün       b +u gj 2 n
bugüne      b +u gj 2 n e

Reflected ordinary words may have more than one pronunciations due to having several morphological analyses:

koyun   koyun    k o j +u n
        koy+un   k +o j u n
        koyu+n   k o j +u n

Some nouns are both geographical names and proper nouns, so there are more than one pronunciations due to stress positions:

Aydın   city          +a j d 1 n
        proper noun   a j d +1 n
        ordinary noun a j d +1 n

Most of the ambiguities in the pronunciations come from the stress marker. Remember we're producing stress positions only for TTS mode, we don't need to generate stress positions for ASR as neural networks are able to recognize Turkish stress pattern anyway. We made a rule-based stress generator module according to these rules, upto the morphological analysis and genre of the word.

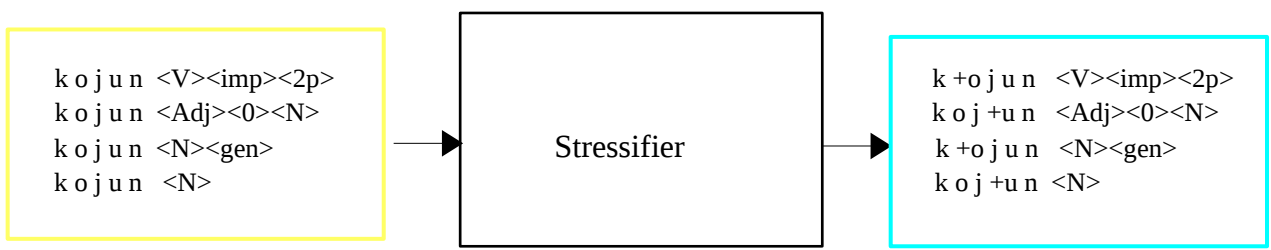

Fig.4. Output of the stress module

## 4. Troubles

There are already several classes of phonetically irregular roots. Ones don't follow final devoicing rule, ones with vowel alterations in case of reflecting, ones with irregular stress pattern….Turkish looks more or less phonetic at the first glance, however as understood from the previous section there are quite many irregularities as well. Although we need to mark dictionaries for these kind of irregularities, irregular roots are well-known and manageable. Real trouble is with foreign named entities and abbreviations. Every day new abbreviations and foreign words get into Turkish written and spoken language. Foreign words are written in several forms; original form, Turkish pronounced form and worst of all original - Turkish mixed form. For instance word *facebook* has several written alternatives: *facebook, feysbuk, feyzbuk, facebuk, feysbuuk*. Foreign named entities are manageable up to some point, abbreviations are much more problematic. We implemented heuristic algorithms to recognize and phoneticizing such words.

### 4.1. Heuristic Stemming

We already emphasized that we used TRMorph for stemming ordinary words. Stemming of Turkish words requires a full morphological analysis as stemming requires identifying possible roots and corresponding suffixes. For instance word "koyun" can be stemmed as:

koyun    sheep
koy+un    your bay
koyu+n    your darkness
koy+un    drop it!

TRMorph and other efficient Turkish stemmers are implemented using finite state tools. Stemming process needs a list of possible roots. Turkish morphological analyzers uses a dictionary of root words with categories noun, adjective, verb etc.
We have no problem with stemming the usual Turkish words. Trouble is stemming foreign, loan words and abbreviations. Turkish language is bombarded with loan and foreign words with the effect of social media and internet.
Foreign words that were borrowed into Turkish , have already became Turkish roots, e.g:

*televizyon* ← *television*
*revision* ← *revision*
*sinema* ← *cinema.*

Such loan words are counted to Turkish roots and exist in list of ordinary roots file, therefore recognized by the morphological analyzer:

*televizyonumdan* → *televizyon+um+dan*

Problem begins with foreign named entities and their reflected forms. We can't write every possible foreign word used in Turkish written language, into dictionary of roots. With the massive effect of social media, hundreds of foreign words and their reflections joins the written language. Main problem is, foreign words might be written in their original form, Turkish pronunciation or half original-half Turkish form. See the different forms of the words "*from my facebook*" and "*in twitter*":

*facebookumdan   feysbukumdan   facebuğumdan*
*facebukumdan   feysbuğumdan*

*twitterda   tivitırda*
*tiviterda   twitırda*

Here's how heuristic stemming works: For recognizing the root boundary, we look for clues. V+voiced consonants *{b, c, d, g} / ğ* + V might indicate a final consonant change, hence a root boundary. We check if the word ends with mos common and characteristic reflectional suffixes: *-sIndAn, -sIndA, -IndAn, -IndA, -dAn, -dA*...See the examples:

*facebuğumdan* → *u+ğ+u* points to a stem boundary. Candidate root is *facebuk*.

*feysbukumdan* → Candidate roots are *feysbuk* and *feysbukum*. Since *-um* is a Turksh suffix and not a typical English digram, most probable root is *feysbuk*.

We generate candidate roots, then we look the roots up in abbreviations and foreign word dictionaries. If root exists in one or both of dictionaries then we generate the corresponding analysis. Currently there are 355 entries in abbreviations dictionary. Foreign words dictionary is cloned from the CMUdict.
If candidate roots are not in any of the dictionaries, we heuristically try to understand if word is foreign, abbreviation or mistyped Turkish. If the candidate root is 2 or 3 letters long, then it's more likely to be an abbreviation. Longer words are discriminated as foreign or mistyped Turkish.
Most loan words of Turkish come from English, we implemented a "does word looks non-Turkish/does word looks English" approach. Lack of vowel harmony points to foreign words. We have a list of common English and Turkish trigrams and tetragrams, as well as Turkish phonetized forms of English counterparts. For instance, *con*, *tha*, *tive* are common sequences of English and Turkish phonetized counterparts are *kon*, *de*, *tiv*. For instance:

f*ace*buk    *ace* is an English trigram and not a Turkish trigram.

f*eys*buk    *eys* is not a Turkish trigram. It's not English either, but it's Turkish pronunciation of the English trigram *ace*.

We used most common and most characteristic suffixes of Turkish for stemming purposes. Common English and Turkish digrams, trigrams and tetragrams are alos used efficiently. Some of the common suffixes are as follows: -lArHndAn, -[s]HndAki, -DAki, -[s]HndAn, -[s]HndA, -[s]HnA, -[s]HnH, -[y]HA, -tA, -tAn, -[n]Hn, -[y]A, -yH, -sH, -HnH, -H, -DAydHm, -DAydH, -DAydHk, -DAydHnHz, -ydH, -HAyken, -HAydH, -lArken, -DH
See the heuristic stemmer module for the implementation details.

### 4.2. Foreign word recognition and phoneticization

As stated in the previous section, there are several problems with recognizing reflected foreign/loan words. We heuristically solved the problem of stemming and recognizing such words in previous section. Now. challenge of generating SAMPA mappings to foreign words comes.
From the view of g2p, *feysbukumdan* and *tivitırda* are completely fine, they're already mapped to Turkish pronunciations. If we feed these words to the phoneticizer module, they'll be correctly mapped to their SAMPA counterparts. But firstly we need to recognize if the word is foreign/loan, abbreviation or just a mistyped Turkish word. We implemented a heuristic foreign word recognition algorithm, but we first need to stem the word.
First we stem the word by the heuristic stemmer. After recognizing the word as foreign/loan/mixed and stemming, we can generate SAMPA mapping of the root. If the root is already Turkish phonetized as *feysbuk*, we directly feed it to the Turkish root phoneticizer. Else, we make a dictionary lookup for the root. If absent in the foreign words dictionary, we feed root to the foreign root phoneticizier module. Foreign root phoneticizier module is rule-based as the rest of the system, built upon phoneticizing rules of most common English trigrams.
Unfortunately trouble doesn't end here. Pronunciation of an English word varies in Turkish society according to educational level and knowledge of English language. In the latter case, one pronounces the word according to the Turkish phonetic rules. Some examples are:

*google*      /g u g 1 5/
              /g o g 1 5/
              /g o g l e/
*gemini*      /gj e m i n i/
              /dZ e m i n i/
*generation*  /dZ e n e r e j S 1 n/
              /gj e n e r a s j o n/

Our rule-based foreign word phoneticizier generates multiple SAMPA mappings, one closer to Turkish pronunciation and one closer to English pronunciation with Turkish mouth. Currently there are 427 entries in the foreign words dictionary. Of course English phonetics is not possible to conquer with a rule-based strategy. Future work involves preparing an English words – Turkish SAMPA transcriptions learning dictionary, then prepare a statistical g2p for English words. Please see the future work section.

**4.3. Abbreviation recognition and phonetization**

There are two types of abbreviations in Turkish, ones that are usually expanded and ones that are not. For instance *ptt*, *bddk* are read letter by letter, while *doç.* and *dr.* are always expanded to their word counterparts, *doçent* and *doktor*. We expand latter class in normalization step, and leave first class as they are. Recognizing foreign/mixed/loan words are much easier than discriminating abbreviations from mistyped Turkish words. Abbreviations follows neither Turkish nor English grapheme statistics. Discriminating short foreign words from abbreviations is also not easy. For instance word *ulm*, looks like both an abbreviation and a foreign word; corresponding pronunciations are:

*ulm*   /u 5 m/           if foreign
*ulm*   /u l e: m e:/
        /u e l e m/       if abbreviation

After doing a heuristic stemming, we are ready to generate SAMPA mappings are abbreviations:

2 letter abbreviations are easy. V-V, C-C, V-C, and C-V abbreviations are usually pronounced letter by letter.

*tr*   /t e: r e:/
*aa*   /a: a:/
*ab*   /a: b e:/

3 letter abbreviations are sometimes quite ambiguous. For instance V-C-C abbreviations might be pronounced as a word or letter by letter, see the following examples:

*akp*   /a: c e: p e:/
*aft*   /a f t/

If there's no vowel in the abbreviations, we pronounce letter by letter:

*stm*   /s e: t e: m e:/
*thy*   /t e: h e: j e:/

C-V-C, V-C-V, C-V-V and C-V-V-C patterns are usually pronounced as words:

*sat*   /s a t/
*itü*   /i t y/
*tüik*  /t y: i c/
*tai*   /t a j i/

V-V-C can be pronounced either way, as a word or letter by letter:

*aal*   /a: a: l e:/
*aal*   /a: a 5/

C-C-V and V-V-C are usually pronounced letter by letter:

*mta*   /m e t e: a:/
*aet*   /a: e: t e:/

Foreign abbreviations follow their original pronunciations:
*mtv* /e m t i: v i:/
*ntv* /e n t i: v i:/
*ai* /e j a j/

Notice that all abbreviation pronunciations follows Turkish phonetics, in the case of two vowels coming together, either we put a slight /j/ sound or a glottal stop in between, depending on the surrounding vowels. Agglutinative nature of Turkish increases ambiguity. For instance, if the abbreviation *thy* is not already in our abbreviations dictionary; *thyde* looks like both a foreign word, a mistyped word and *thy+de*(*at THY*). Foreign words with endings similar to Turkish suffixes doesn't help either. *aidan* might be the English proper noun or *ai+dAn* (*from AI*). Reflected abbreviations looks like both typos and foreign words.

As seen, neither recognizing nor phoneticizing abbreviations is not easy. As previously said, abbreviations don't follow neither Turkish nor English grapheme statistics. Hence comparisons with Turkish and English trigrams didn't lead to any improvement.

We implemented a heuristic stemming algorithm, described in foreign word recognition section. After stemming, if the candidate root is 2 or 3 characters long, then it's most probably an abbreviation. Then we make a dictionary lookup for the candidate root in our abbreviations dictionary. If root pronunciation is not present in we feed the root to the heuristic abbreviation pronunciation generator module. Currently there are 355 entries in abbreviations dictionary.

## 5. Some implementation details

As we already emphasized, we used TRmorph of Çağrı Çöltekin for morphological anlaysis. TRmorph is being developed with foma. Foma is a compiler, programming language, and C library for constructing finite-state automata and transducers for various uses.

We made our implementation in Python due to ease of string processing. We also used foma Python wrapper due to use of TRmorph.

See the https://github.com/DuyguA for our implementation.

We also implemented a statistical g2p as well, see the above GitHub repository for the variant made with Phonetisaurus and OpenFst.

## 6. Future work

Future work includes making a statistical g2p for English words. We'll prepare a "English words with Turkish mouth" dictionary, then use Phonetisaurus on it to produce a statistical g2p.